\documentclass[10pt,twocolumn,letterpaper]{article}
\usepackage[accsupp]{axessibility}
\usepackage{color}
\usepackage{cvpr}
\usepackage{times}
\usepackage{epsfig}
\usepackage{graphicx}
\usepackage{amsmath}
\usepackage{amssymb}
\usepackage{multirow}
\usepackage{multicol}
\usepackage{subfigure}

\usepackage[pagebackref=true,breaklinks=true,letterpaper=true,colorlinks,bookmarks=false]{hyperref}

\cvprfinalcopy 

\ifcvprfinal\fi
\begin{document}

\title{Deep Image-based Illumination Harmonization}

\author{Zhongyun Bao\textsuperscript{1},
Chengjiang Long\textsuperscript{2}\thanks{This work was co-supervised by Chengjiang Long and Chunxia Xiao.},  
Gang Fu\textsuperscript{1},
Daquan Liu\textsuperscript{1},
Yuanzhen Li\textsuperscript{1},
Jiaming Wu\textsuperscript{1},
Chunxia Xiao\textsuperscript{1}$^{\tt\small*}$\thanks{Corresponding author.}\\
\textsuperscript{1}{School of Computer Science, Wuhan University, Wuhan, Hubei, China}\\
\textsuperscript{2}{ Meta Reality Labs, Burlingame, CA, USA}\\
{\tt\small clong1@fb.com, xyzgfu@gmail.com, \{zhongyunbao, daquanliu, cxxiao\}@whu.edu.cn}
}

\maketitle
\thispagestyle{empty}

\begin{abstract}
Integrating a foreground object into a background scene with illumination harmonization is an important but challenging task in computer vision and augmented reality community. Existing methods mainly focus on foreground and background appearance consistency or the foreground object shadow generation, which rarely consider global appearance and illumination harmonization. In this paper, we formulate seamless illumination harmonization as an illumination exchange and aggregation problem. Specifically, we firstly apply a physically-based rendering method to construct a large-scale, high-quality dataset (named IH) for our task, which contains various types of foreground objects and background scenes with different lighting conditions. Then, we propose a deep image-based illumination harmonization GAN framework named DIH-GAN, which makes full use of a multi-scale attention mechanism and illumination exchange strategy to directly infer mapping relationship between the inserted foreground object and the corresponding background scene. Meanwhile, we also use adversarial learning strategy to further refine the illumination harmonization result. Our method can not only achieve harmonious appearance and illumination for the foreground object but also can generate compelling shadow cast by the foreground object. Comprehensive experiments on both our IH dataset and real-world images show that our proposed DIH-GAN provides a practical and effective solution for image-based object illumination harmonization editing, and validate the superiority of our method against state-of-the-art methods. Our IH dataset is available at \url{https://github.com/zhongyunbao/Dataset}.
\end{abstract}

\begin{figure}[htb]
\subfigure[]{\includegraphics[width=0.10\linewidth, height=0.16\linewidth]{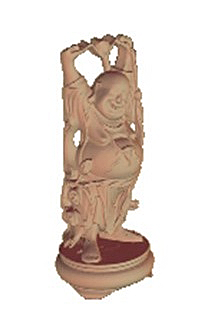}\vspace{0pt}}
\subfigure[]{\includegraphics[width=0.28\linewidth, height=0.19\linewidth]{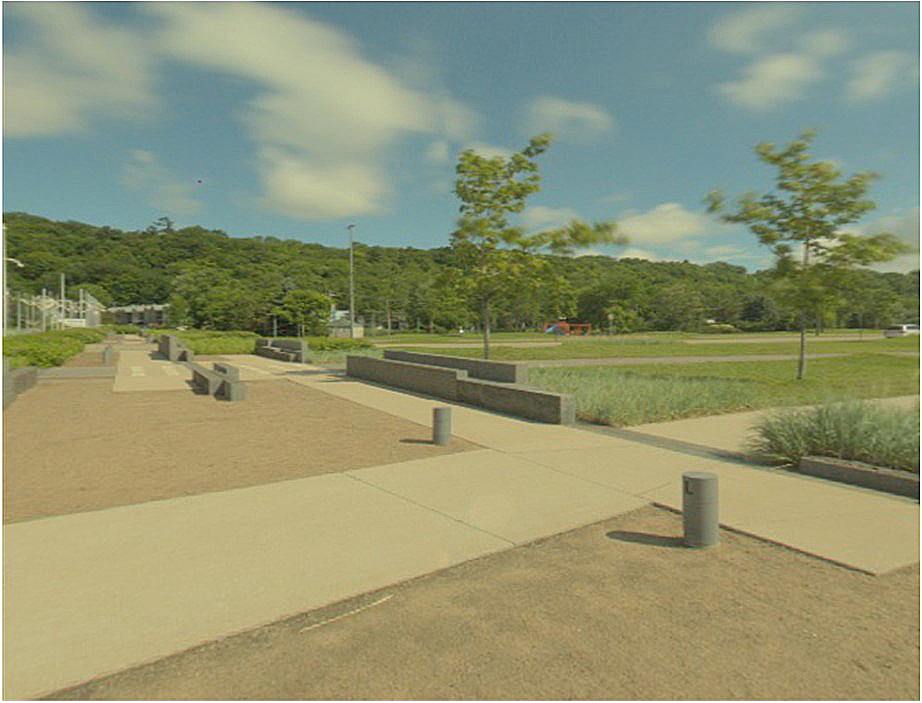}\vspace{0pt}}
\subfigure[]{\includegraphics[width=0.28\linewidth, height=0.19\linewidth]{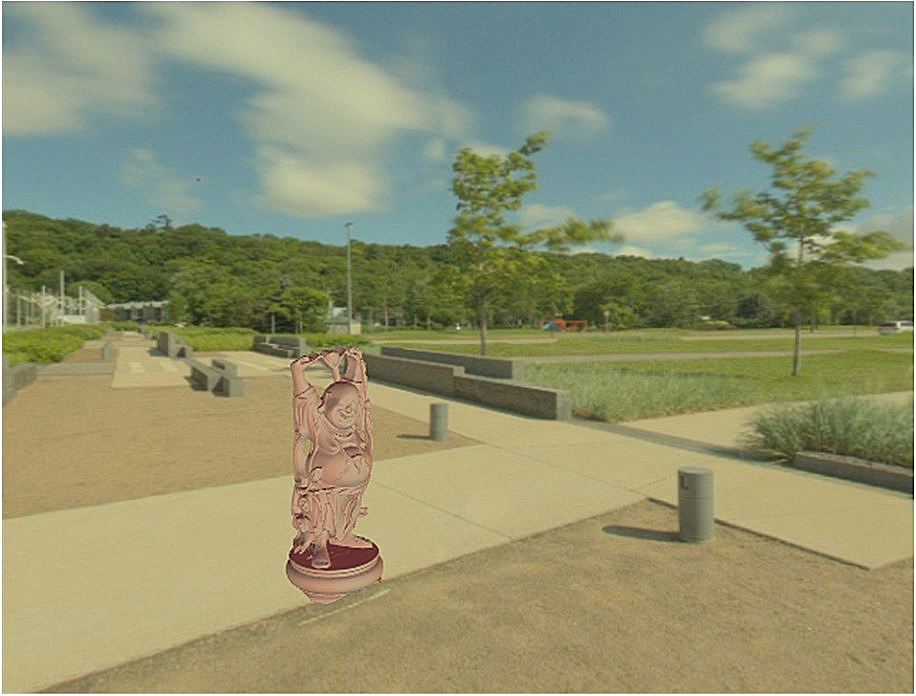}\vspace{0pt}}
\subfigure[]{\includegraphics[width=0.28\linewidth, height=0.19\linewidth]{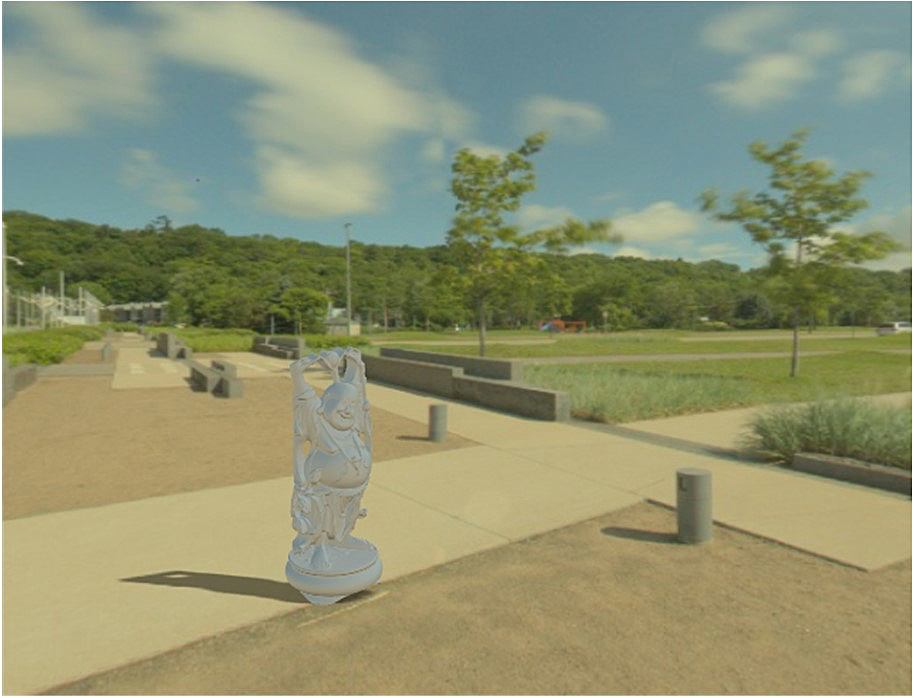}\vspace{0pt}}
\caption{Illumination editing for an inserted object in a single image. (a) Foreground object imagr  with a illumination condition. (b) Background image  with a new illumination condition. (c) Naive composite image. (d) Illumination harmonized image.}
\label{fig:introduction}
\end{figure}

\section{Introduction}
 As a part of scene editing, editing illumination for inserted object to achieve scene illumination harmonization is really important in computer vision~\cite{chen2021canet, fu2021multi, fu2020learning, yu2021luminance} and  augmented reality (AR) as the unsatisfactory illumination harmonization greatly affect the user sense of reality. One example is illustrated in Figure~\ref{fig:introduction}. It is still difficulty to achieve a satisfactory result for illumination editing even by a experienced professional retoucher. Apparently, it is very challenging to automatically edit illumination harmonization without any human intervention.

Some prior efforts have been made to solve this challenging task. In particular, 
Karsch \textit{et al}.~\cite{Karsch2014Automatic} presented an image editing system that supports drag-and-drop 3D object insertion, and Liao \textit{et al}.~\cite{2015An,2018An} proposed an approximate shading model for image-based object modeling and insertion. Although these methods produce perceptually convincing results, their performances highly depend on the quality of the estimated geometry, shading, albedo and material properties. However, in some cases, any errors or inaccurate estimation in either geometry, illumination, or materials may result in unappealing editing effects. 

Such a shortcoming strongly motivates us to explore a deep learning-based method to directly learn mapping relationship between the inserted image-based foreground object and the real-world scene, and achieve scene illumination harmonization without any explicit inverse rendering (recovering 3D geometry, illumination, albedo and material). 
Obviously, a dataset with a lot of training image pairs of composite images without illumination harmonization and corresponding ground truth with illumination harmonization is strongly desired for the training purpose. 

However, existing datasets like iHarmony4 \cite{2020DoveNet}, shadow-AR dataset \cite{2020ARShadowGAN}, HVIDIT dataset\cite{guo2021intrinsic}, \textit{etc}., mainly focus on foreground object appearance or the foreground object shadow,  what rarely consider global appearance and illumination harmonization. The dataset \cite{2021Adversarial} considers both appearance and shadow of inserted foreground object, it contains only two types of foreground objects: car and person, which is not only unavailable, but also severely limits the generalization and robustness of the illumination harmonization task. 

\begin{figure*}[ht]
\vspace{-0.75cm}
\subfigure[]{\includegraphics[width=0.04\linewidth, height=0.095\linewidth]{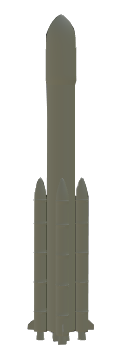}}
  \hspace{-3pt}
\subfigure[]{\includegraphics[width=0.14\linewidth, height=0.095\linewidth]{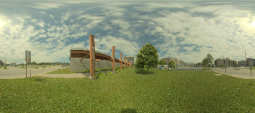}} 
  \hspace{-3.5pt}
\subfigure[]{\includegraphics[width=0.11\linewidth, height=0.095\linewidth]{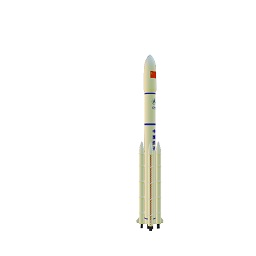}}
  \hspace{-4pt}
\subfigure[]{\includegraphics[width=0.14\linewidth, height=0.095\linewidth]{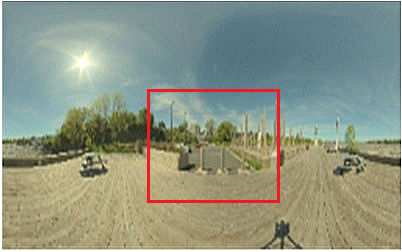}} 
  \hspace{-3pt}
\subfigure[]{\includegraphics[width=0.11\linewidth, height=0.095\linewidth]{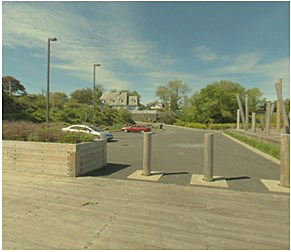}}
  \hspace{-4pt}
\subfigure[]{\includegraphics[width=0.11\linewidth, height=0.095\linewidth]{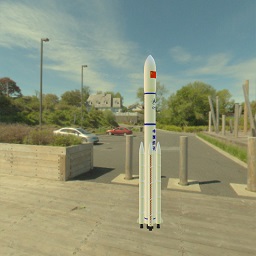}}
\hspace{-4pt}
\subfigure[]{\includegraphics[width=0.11\linewidth, height=0.095\linewidth]{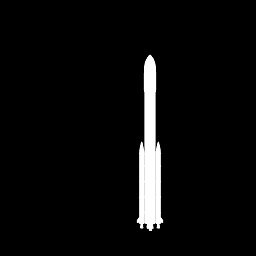}}
  \hspace{-4pt}
\subfigure[]{\includegraphics[width=0.11\linewidth, height=0.095\linewidth]{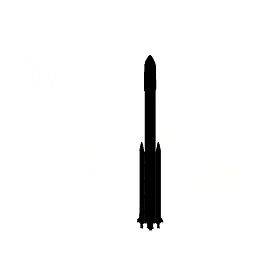}}
  \hspace{-4pt}
\subfigure[]{\includegraphics[width=0.11\linewidth, height=0.095\linewidth]{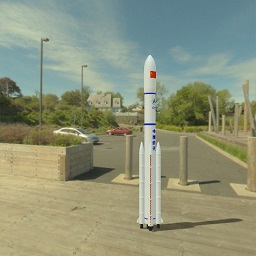}}
\caption{The illustration of a synthesized illumination harmonization image generation process. Given a 3D object with texture (a), we first apply a panorama illumination corresponding with (b) to render the 3D object and get an image-based object (c). Then we paste the image-based object into a background image (e) cropped from (d) directly without illumination adjustment. In this way, along with an object mask (g) and a background mask (h), we get a naive composite image (f) with illumination and shadow inconsistency between object and its surrounding. With the background illumination map (d), we then use Blender to synthesize the illumination harmonization image (i) and take it as the ground-truth for supervised learning. 
Note we consider (f), (g) and (h) as an input triplet, and (b), (d) and (i) as a ground truth triplet.
The input triplet and corresponding ground truth triplet are treated as a six-tuple in our dataset.
\textit{Better view in electronic version}.}
\label{fig:data}
\end{figure*}

In this work, we first construct a large-scale, high-quality synthesized dataset named IH dataset for the object illumination editing task. To build our dataset, we first collect HDR panoramas to capture background images and illumination information from Laval's HDR dataset \cite{Gardner2017Learning, gardner2019deep} and the Internet, which are taken in various indoor and outdoor real-world scenes.
Therefore, the scenes in our dataset are general and challenging. Besides, we also collect 60 3D object models with considerably different shapes and postures used as the foreground objects of our composite images. In general, our dataset finally contains 89,898 six-tuples in total, each with one input triplet (\textit{i.e.}, a naive composite image, and the corresponding object mask and background mask), and another ground truth triplet (\textit{i.e.}, a foreground object illumination map, a background illumination map, and a final illumination harmonization image). See Figure~\ref{fig:data} for a six-tuples example.


Regarding to the deep learning model, we propose a novel learning-based scene illumination harmonization GAN framework named DIH-GAN, as shown in Figure~\ref{fig:SI-GAN}, which incorporates both spatial attention learning~\cite{2020An, Islam:CVPR2020, Islam:AAAI2021, Hu:TIP2021, Vasu:WACV2020} 
and adversarial learning ~\cite{goodfellow2014generative,2017Wasserstein} to make the illumination of foreground  compatible with background.  Our DIH-GAN takes a naive composite image with shadow-free object as well as inserted object mask as input, and makes full use of multi-scale attention mechanisms and adversarial learning to directly infer mapping relationship between the inserted foreground object and the corresponding background scene. Besides, we propose an illumination exchange mechanism to edit object illumination and directly achieve seamless illumination integration between the foreground object and the background of composite image, which makes the synthesized image more harmonious and realistic. Note that, our proposed multi-scale attention mechanism and feature exchange mechanism play a key role, which can avoid the complicated inverse rendering process and directly generate reasonable illumination harmonized results. 

Our main contributions are summarized as follows:
\begin{itemize}
  \vspace{-0.3cm}
  \setlength{\itemsep}{0pt}
  \setlength{\parsep}{0pt}
  \setlength{\parskip}{0pt}
\item We construct the first large-scale, high-quality image illumination harmonization dataset IH, which consists of 89,898 image six-tuples with a diversity of real-world background scenes and 3D object models. 
\item We propose a novel deep learning-based scene illumination harmonization GAN framework named DIH-GAN, which is a multi-task collaborative network and can directly perform illumination harmonization editing for the inserted object without explicit inverse rendering.
\item Extensive experiments show that the proposed DIH-GAN can effectively achieve high-quality image illumination harmonization and significantly outperform existing state-of-the-art methods. 
\end{itemize}


\section{Related work}

\noindent{\bfseries{Object illumination editing.}}
Traditional object illumination editing methods mainly concentrated on estimating the scene geometry, illumination and surface reflectance to edit the object. Previous methods \cite{Jean2007Photo,karsch2011rendering} have shown that coarse estimation of scene geometry, reflectance properties, illumination, and camera parameters work well for many image editing tasks. These methods require a user to model the scene geometry and illumination. The method \cite{2013Intrinsic} not only recovers shape, surface albedo and illumination for entire scenes, but also requires a coarse input depth map, while this method is not directly suitable for illuminating inserted object. Karsch \textit{et al}. \cite{Karsch2014Automatic} presented a fully automatic method for recovering a comprehensive 3D scene model (geometry, illumination, diffuse albedo and camera parameters) from a single low dynamic range photograph. Liao {\em et al.} \cite{2018An} presented an object relighting system that supports image-based relighting, although this method achieves impressive result, it still needs to reshape the object and model the scene. 

These methods depend on the physical modeling of object and scene information, and inaccurate reconstruction results will lead to poor results. In contrast, our method automatically edits the object illumination, directly generates harmonized illumination results without complicated inverse rendering, and thus produces better visual effects.

\noindent{\bfseries{Shadow generation.}} Recently, with the breakthrough in adversarial learning, generative adversarial network (GAN) \cite{goodfellow2014generative,2017Wasserstein,2014Conditional} have been successfully applied to shadow detection, removal and generation \cite{wang2018stacked,2020ARGAN, 2019Mask, 2019ShadowGAN, 2020ARShadowGAN}. For shadow generation, Liu {\em et al.} \cite{2020ARShadowGAN} proposed an ARShadowGAN model, which is able to directly model the mapping relation between the shadow of the foreground object and the corresponding real-world environment based on their constructed dataset. Similar to this method, our method also aims to generate the object shadow without explicit estimation of 3D geometric information. Besides that, our method considers the shading of the object itself. We not only realize reasonable object shadow generation with the similar effect as Liu {\em et al.}'s \cite{2020ARShadowGAN} method, but also edit the object illumination to achieve overall scene illumination harmonization.   

\noindent{\bfseries{Image-to-image translation.}}
Image-to-Image translation is to map an input image to a corresponding output image. It has been widely used in various tasks, including super-resolution \cite{2016Accurate, 2016Photo}, image quality restoration \cite{mao2016image,2014Deep}, image harmonization \cite{isola2017image,2020DoveNet,ling2021region,guo2021intrinsic}, and so on. It is worth mentioning that Cong \textit{et al}. \cite{2020DoveNet} proposed a novel domain verification discriminator, with the insight that the foreground needs to be translated to the same domain as the background for image harmonization, but neglect to explicitly transform the foreground features in the generator. Recently, Ling \textit{et al}. \cite{ling2021region} treated image harmonization as a style transfer problem to explicitly formulates the visual style from the background and adaptively applies them to the foreground, Guo \textit{et al}. \cite{guo2021intrinsic} modeled image harmonization based on intrinsic image theory.

These methods all focus on the illumination of foreground object and do not consider object shadow generation task. Different from all existing methods, our task takes into account both illuminating the object and generating the cast shadow of the object, and achieves illumination harmonization for the whole scene.

\section{Our IH Dataset}
\label{sed:ourDataset}

The construction process of the IH dataset includes three steps: (1) collecting images and 3D models, (2) filtering background images, and (3)
rendering and composition. In the following, we will describe these steps in details.

\noindent{\textbf{Collecting images and 3D models.}} We first collect all images from the Laval's HDR panorama dataset\footnote{The author Zhongyun Bao signed the license and produced all the experimental results in this paper. Meta did not have access to the Laval's HDR panorama dataset.} \cite{Gardner2017Learning, gardner2019deep}, and capture 2,686 HDR panorama images from the Internet with a diversity of real-world scenes. For each panorama image, we extract 8 limited field-of-view crops to produce the background images of composite image, and also use the corresponding panorama images centering on the crops as the illumination to render ground truth results. We initially obtain 22,256 background images with the corresponding illumination map in total. Also, we collect 60 3D models from the website (\url{https://laozicloud.com})\footnote{The author Zhongyun Bao purchased the 3D models for non-commercial research purpose only and produced the experimental results in this paper. Meta did not have access to these data.} as inserted objects, which contain various types of objects, such as \textit{bunny},  \textit{person}, \textit{lucy} \textit{etc}. 

\noindent{\bfseries{Filtering background images.}} To ensure the quality of the dataset for our object illumination editing task, we further filter out the following three kinds of images: (1) without obvious or natural-looking illumination, (2) without a reasonable place to insert virtual object, and (3) with inconspicuous or no shadows. By this way, we finally obtain 12,253 remaining background images.

\noindent{\bfseries{Rendering and compositing.}} With collected 3D models, background images, and the corresponding panorama maps, the ground truth object relighting images (see Figure~\ref{fig:data} (i)) are rendered using Blender. 
Specifically, we first specify a plane at the bottom of the inserted object for casting shadows, then embed the 3D object into the cropped background image, and finally use the corresponding panorama map to render the illumination of the object to produce the final result.
Note that, due to the corresponding backgrounds are real-world 2D scene images, we use Photoshop to manually annotate each foreground object in our dataset to  obtain accurate masks.

\begin{figure*}[t]
\centering
\includegraphics[width=0.98\textwidth]{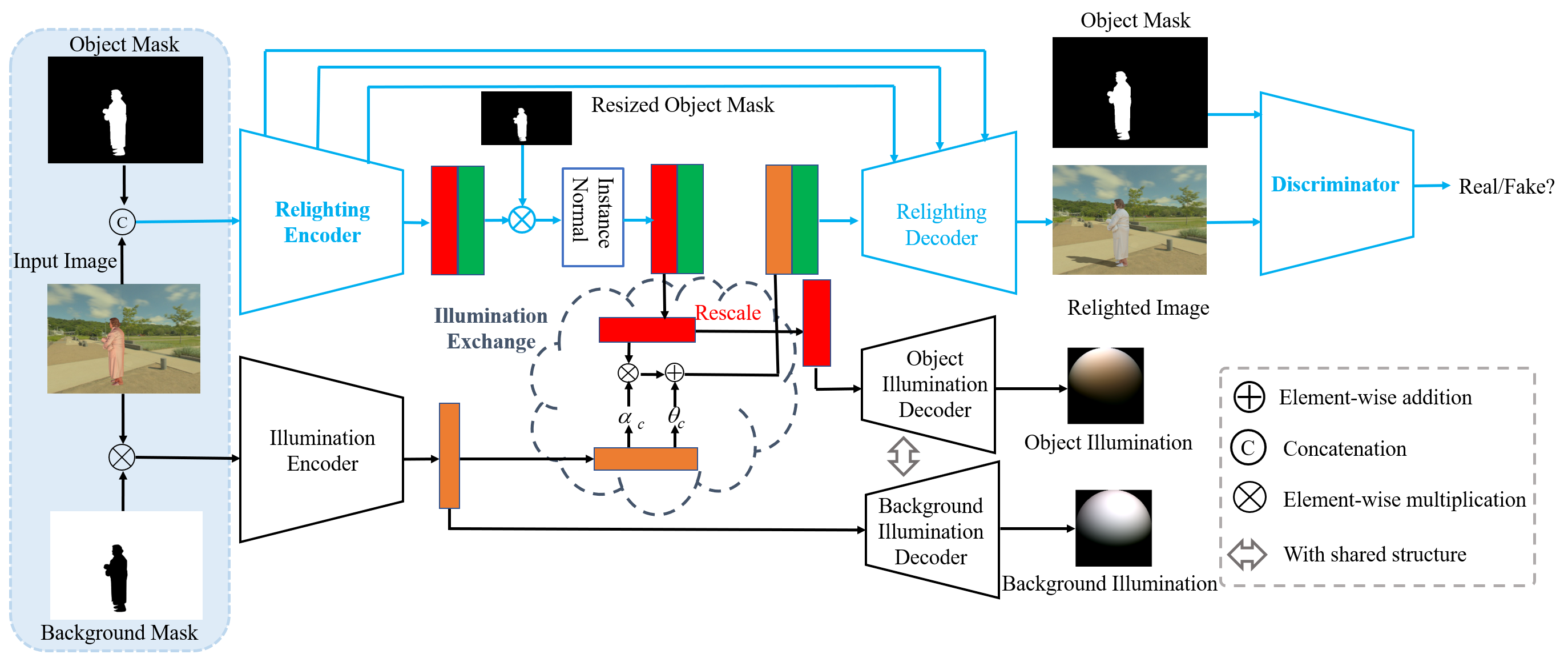}
\caption{The overview of our proposed DIH-GAN. Given an input image with inserted object and the corresponding object mask and background mask, the generator of our DIH-GAN can generate the relighting image (R-Network marked in blue) and predict both object illumination and background illumination (I-Network marked in black), and the discriminator can distinguish whether the generated relighting image is real or fake. 
The Illumination exchange mechanism between the R-Network and the I-Network realizes the conversion of illumination information between the scene and the object.}
\label{fig:SI-GAN}
\end{figure*}
In the construction process, we use 60 virtual models with different pose configures using the pipeline shown in Figure~\ref{fig:data} to construct our dataset based on different background images, and we produce 169,672 synthesized ground truth illumination harmonization images in total.  Moreover, to improve the training efficiency, we only use 89,898 six-tuples to train the our network in final. Each six-tuple consists of two triplets. One triplet as input data includes a naive composite image without illumination adjustment, and the corresponding object mask and background mask. The other one as ground-truth data includes a synthesized illumination harmonization image, one object illumination and one background illumination ground-truths. A visual six-tuple example is shown in Figure~\ref{fig:data}. Refer to the supplementary for more details of the dataset analysis.

\section{Proposed Method}
\label{sec:Method}

Our goal is to train a GAN that takes a naive composite image $\hat{Y}$  with inharmonious illumination, corresponding the object mask, background mask and corresponding target illumination as input, and directly generate the corresponding scene illumination harmonized image $\bar{Y}$. To achieve this goal, we propose a novel framework called DIH-GAN, of which the generator is a multi-task parallel network composed of two networks, \ie, Relighting Network (R-Network) and Illumination Network (I-Network) to handle object and illumination separately. See Figure~\ref{fig:SI-GAN}. 



\subsection{ Generator}
As shown in Figure~\ref{fig:SI-GAN}, the generator of our DIH-GAN contains two parallel branch networks, {\em i.e.}, R-Network and I-Network. 
R-Network learns the overall features of the input image and I-Network predicts the object and background illumination. They work collaboratively to complete the task by illumination exchange mechanism.


{\noindent\bf Relighting Encoder}. For the U-Net \cite{ronneberger2015u} like R-Network, there are five down-sampling blocks in encoder and each down-sampling block consists of a residual block with 3 consecutive convolutions, instance normalization and ReLU operation and halves the feature map with an average pooling operation. Each down-sampling block is followed by a multi-scale attention block which guides the network to infer the object shadow and generates the refinement feature maps. Note that we design such a multi-scale attention mechanism for two purposes: (1) to adaptively extract reliable multi-scale features and overcome the scale-variation across the image to assign larger weights to areas of interest for refinement; and (2) to guide the generation of shadows of the inserted objects by paying attention to real shadows and corresponding occluders in the scene.
\begin{figure}[] 
\centering
\includegraphics[width=0.48\textwidth]{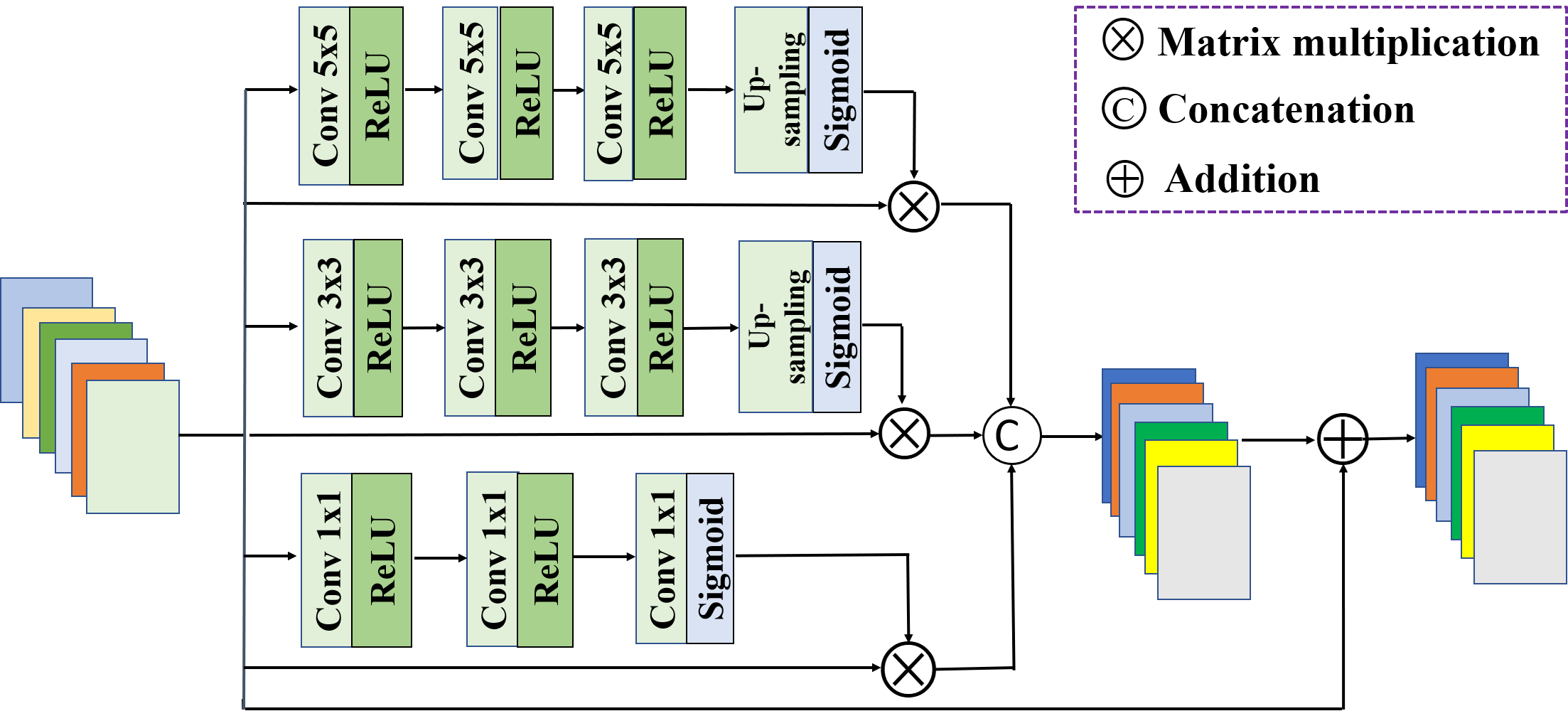}
\caption{Illustration of our multi-scale attention mechanism.}
\label{fig:MSA}
\end{figure}

As shown in Figure~\ref{fig:MSA}, the multi-scale attention block has three types convolution layers with three different kernel sizes, $1\times1$, $3\times3$, $5\times5$, to extract features in different scales. Specially, for the input feature map, the multi-scale attention block first extracts features using two $1\times1$ convolution layers with crossing channels and squeezing features, two $3\times3$ convolution layers and two $5\times5$ convolution layers to generate feature maps. Note that for $3\times3$ and $5\times5$ convolution, the feature map size of each channel has been changed and therefore we apply an up-sampling layer to recover the original size before feeding the feature map into the Sigmoid function to produce attention map. We separately conduct an element-wise multiplication on the input feature and the attention map at each scale to produce attended feature maps, which are then concatenated at channel-wise together and fed into a $1\times1$ convolution layer to recover the same channel number with the origin input feature. We apply a residual structure \cite{He2016Deep} to combine it with the origin input feature map together as final output. This residual mechanism not only accelerates the convergence speed but also correct image details such as border artifacts.

The final output features of the encoder include the illumination features $F_{illu}$ and non-illumination features $F_{noillu}$ of global image. This feature separation is enforced by the no-illumination loss $\mathcal{L}_{noillu}$ (see Eq.~(\ref{eqn:noillum_loss})).

{\noindent\bf Illumination Encoder.} For the I-Network, the encoder has a similar structure to the one of the R-Network and takes the result of multiplying the input image and background mask as input to extract the illumination feature of background. The background illumination feature is then exchanged with the object illumination feature of R-Network by the illumination exchange mechanism. Note that the output of illumination encoder have two functions: the background illumination features are used in combination with object non-illumination features $F_{noillu}$ of R-Network encoder to produce the illumination harmonized image in the decoder of R-Network; the background illumination features are also fed to the decoder of I-Network to predict the corresponding background illumination information through supervised learning. 

{\noindent\bf Illumination Exchange Mechanism}. After obtaining the features extracted by the two encoders, inspired by \cite{ling2021region}, we use the background illumination features in the I-Network to guide the foreground object illumination features in the R-Network, and exchange them for the input of the decoders. 

To specify, the two sub-networks work together through the illumination exchange mechanism. It is worth mentioning that at the bottleneck feature of the R-Network, we perform multiplication operation on it with the object mask of the corresponding size, and get the normalized foreground object features $F^{obj}$ by using IN \cite{ulyanov2016instance}. This treatment is able to better realize the exchange of object illumination and background illumination, and achieve the illumination harmonization task. 

The normalized foreground object feature $F^{obj}$ can be divided into two parts: non-illumination features $F_{noillu}^{obj}$ which is independent of illumination feature $F_{illu}^{obj}$. The illumination feature $F_{illu}^{obj}$ have two functions: one is to be cropped by the resized object mask, rescaled to a larger size and then fed into the object illumination decoder of the I-Network to predict the object illumination. The other is to be affined by learned scale and bias from the background illumination features $F_{illu}^{bg}$ extracted by the I-Network encoder, and then the affined features are concatenated with $F_{noillu}^{obj}$ and fed into the R-Network decoder to generate the realistic Illumination harmonization image. The affine result is computed by:

\begin{equation}
\label{affine}
\mathcal F_{affine} = \alpha_{c} F_{illu}^{obj} + \theta_{c},
\end{equation}
where $\theta_{c}$ and $\alpha_{c}$ are the mean
and standard deviation of the activations of the background illumination features in channel c:
\begin{equation}
\label{A}
\mathcal{\theta}_{c} = \frac{1}{N_{bg}}\sum_{h,w} F_{illu}^{bg},
\end{equation}

\begin{equation}
\label{B}
\mathcal{\alpha}_{c} =\sqrt{\frac{1}{N_{bg}}\sum_{h,w} (F_{illu}^{bg} - {\theta}_{c})^2},
\end{equation}
where $N_{bg}$ is the total number of pixels of background illumination, $h, w$ denote the height and width of features, respectively.

In the whole illumination harmonization task, we have supervised constraints on the relighting image, the non-illumination and illumination feature of the object, and the background illumination feature, respectively, which improves the harmonization accuracy of the relighting image. 

{\noindent\bf Relighting Decoder.} 
The decoder in the R-Network consists of five up-sampling layers. Each up-sampling layer doubles the feature map by nearest interpolation followed by consecutive dilated convolution, instance normalization and ReLU operations. The last feature map is activated by a sigmoid function. The R-network concatenates down-up sampling layers by skip connections. 

{\noindent\bf Object/Background Illumination Decoders.} Following~\cite{yu2019inverserendernet}, the decoder of the I-Network is to predict the illumination. In this paper, we use the shared structure for both object illumination and background illumination decoders. 

\subsection{Discriminator}
The discriminator of DIH-GAN is designed to help the R-Network accelerate convergence and generate a plausible harmonized image. Following Patch-GAN \cite{isola2017image}, our discriminator consists of six consecutive convolutional layers. Each convolutional layer contains convolution, instance normalization and ReLU operations. We use Sigmoid function to activate last feature map produced by a convolution, and perform a global average pooling operation on the activated feature map to obtain the final output of the discriminator. 

\subsection{Loss functions}
The total loss $\mathcal{L}_{total}$ is formulated with an illumination loss $\mathcal{L}_{illu}$, a non-illumination loss $\mathcal{L}_{noillu}$, a perceptual loss $\mathcal{L}_{per}$, an adversarial loss $\mathcal{L}_{adv}$ , and a classical ${L}_{1}$-normal reconstruction loss $\mathcal{L}_{Recons}$ as follows:
\begin{equation}
     \begin{split}
         \mathcal{L}_{total} = &\beta_{1}\mathcal{L}_{illu} + \beta_{2}\mathcal{L}_{noillu} + \beta_{3}\mathcal{L}_{per} \\
        & + \beta_{4}\mathcal{L}_{adv} + \mathcal{L}_{Recons},
      \end{split}
\end{equation}

where $\beta_{1}$, $\beta_{2}$ , $\beta_{3}$, $\beta_{4}$ are weighting parameters controlling the influence of each term.

{\bfseries{ Illumination loss}} $\mathcal{L}_{illu}$ is the element-wise illumination loss between the generated illumination and the corresponding ground truth, {\em i.e.},
\begin{equation}
\begin{split}
&\mathcal{L}_{illu} = \| Y_{OI} - \bar{Y}_{OI}\|_{2}^{2} + \| Y_{BI} - \bar{Y}_{BI}\|_{2}^{2}, 
\end{split}
\label{eqn:illum_loss}
\end{equation}
where $\bar{Y}_{OI}$ and $\bar{Y}_{BI}$ 
represent the output foreground object illumination image, background illumination image, and relighting image 
respectively. ${Y}_{OI}$ and ${Y}_{BI}$
are their corresponding ground truth images.

{\bfseries{Non-illumination feature loss}} $\mathcal{L}_{nonillu}$ is introduced to enforce non-illumination feature matching to improve the accuracy of the object relighting image. 
According to \cite {zhou2019deep} and Retinex theory \cite{land-1971-light-retin-theor}, reflectance is the inherent physical property of object, independent of illumination. Therefore, we expect that the same object under different illumination conditions have the same non-illumination (\textit{i.e.}, reflectance) features,
\begin{equation}
\begin{split}
\mathcal{L}_{nonillu} = (F_{nonillu}^1 - F_{nonillu}^2)^{2}/{N_{nonillu}},
\end{split}
\label{eqn:noillum_loss}
\end{equation}
where $F_{nonillu}^1$ and $F_{nonillu}^2$ are non-illumination features of the same insert object under the different illumination conditions, and $N_{nonillu}$ is the number of elements in $F_{nonillu}$.

{\bfseries{ Perceptual loss}} $\mathcal{L}_{per}$ \cite{2016Perceptual} is used to measure the semantic difference between the generated image and the ground truth. Following \cite{2020ARShadowGAN}, we use a VGG-16 model \cite{2014Very} pre-trained on ImageNet dataset \cite{2009ImageNet} to extract feature and choose the first 10 VGG16 layers to compute feature map. $\mathcal{L}_{per}$ is defined as:
\begin{equation}
     \begin{split}
         \mathcal{L}_{per} = &\text{MSE}(V_{Y_{FI}}, V_{\bar{Y}_{FI}}) + \text{MSE}(V_{Y_{BI}}, V_{\bar{Y}_{BI}})\\
        & + \text{MSE}(V_{Y_{R}}, V_{\bar{Y}_{R}}),
      \end{split}
\end{equation}
where $\text{MSE}$ is the mean squared error, and $V_{i} = \text{VGG}(i)$ is the extracted feature map.

{\bfseries{Adversarial loss}} $\mathcal{L}_{adv}$ is utilized to describe the competition between the generator and the discriminator as:
\begin{equation}
\label{loss:L_GAN}
\mathcal{L}_{adv} = \log(\mathbf{D}(x, m, Y)) + \log(1 - \mathbf{D}(x, m, \bar{Y})),
\end{equation}
where $\mathbf{D}(\cdot)$ is the probability that the image is ``real". $x$ is the input image and $m$ is the corresponding mask, $\bar{Y}$ is the output of the generator of DIH-GAN, and $Y$ is the ground-truth. The discriminator tries to maximize $\mathcal{L}_{adv}$ while the generator tries to minimize it.
\begin{figure*}[ht!]
\vspace{-0.5cm}
\centering
  \includegraphics[height=0.073\textwidth, width=0.04\textwidth]{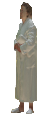} 
  \includegraphics[height=0.093\textwidth,width=0.128\linewidth]{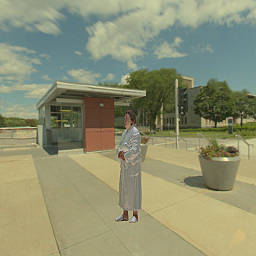} 
  \includegraphics[height=0.093\textwidth,width=0.128\linewidth]{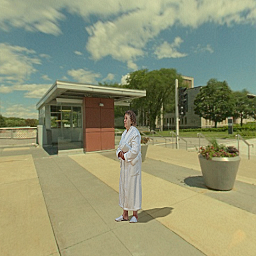}  
  \includegraphics[height=0.093\textwidth,width=0.128\linewidth]{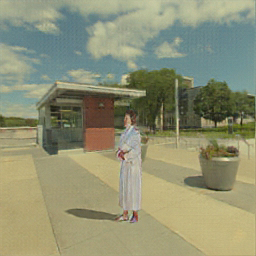}  
  \includegraphics[height=0.093\textwidth,width=0.128\linewidth]{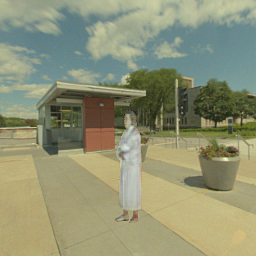}
  \includegraphics[height=0.093\textwidth,width=0.128\linewidth]{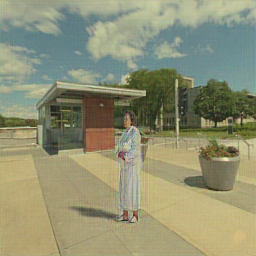}
  \includegraphics[height=0.093\textwidth,width=0.128\linewidth]{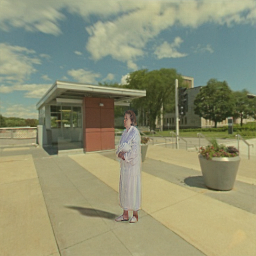}
  \includegraphics[height=0.093\textwidth,width=0.128\linewidth]{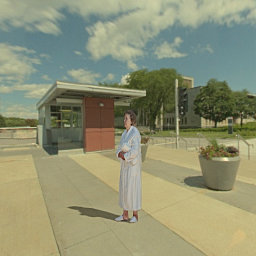} \\
  \vspace{-0.3pt}
  \includegraphics[height=0.073\textwidth, width=0.04\textwidth]{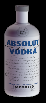}
  \includegraphics[height=0.093\textwidth,width=0.128\linewidth]{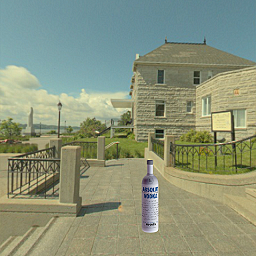}
  \includegraphics[height=0.093\textwidth,width=0.128\linewidth]{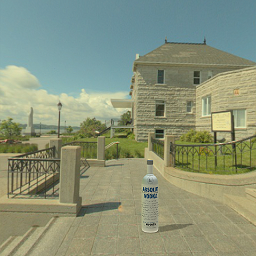} 
  \includegraphics[height=0.093\textwidth,width=0.128\linewidth]{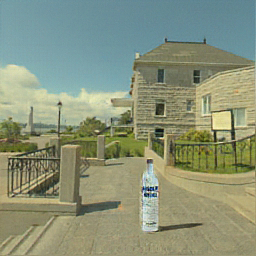} 
  \includegraphics[height=0.093\textwidth,width=0.128\linewidth]{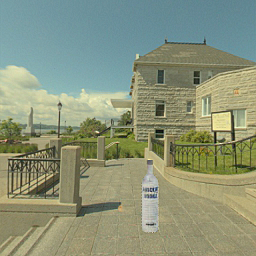} 
  \includegraphics[height=0.093\textwidth,width=0.128\linewidth]{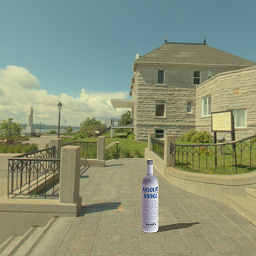}
  \includegraphics[height=0.093\textwidth,width=0.128\linewidth]{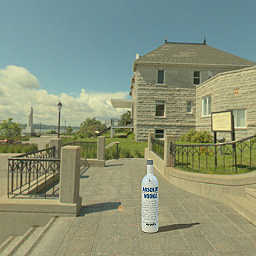} 
  \includegraphics[height=0.093\textwidth,width=0.128\linewidth]{Three_GT_60.png} \\  
  \vspace{-5.pt}
  \subfigure[]{\includegraphics[height=0.063\textwidth, width=0.04\textwidth]{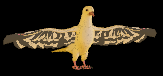}}
  \subfigure[]{\includegraphics[height=0.093\textwidth,width=0.128\linewidth]{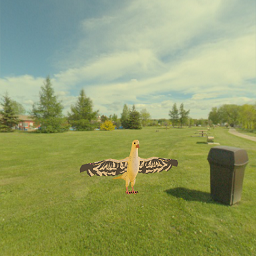}} 
  \subfigure[]{\includegraphics[height=0.093\textwidth,width=0.128\linewidth]{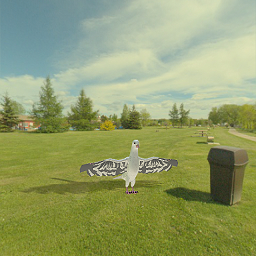}} 
  \subfigure[]{\includegraphics[height=0.093\textwidth,width=0.128\linewidth]{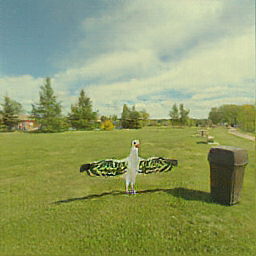}}
  \subfigure[]{\includegraphics[height=0.093\textwidth,width=0.128\linewidth]{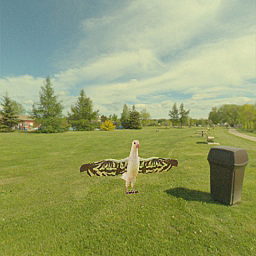}} 
  \subfigure[]{\includegraphics[height=0.093\textwidth,width=0.128\linewidth]{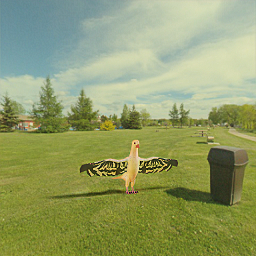}}
  \subfigure[]{\includegraphics[height=0.093\textwidth,width=0.128\linewidth]{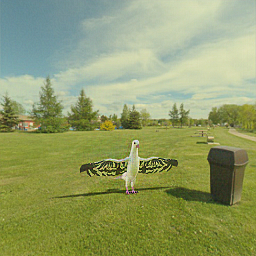}}  
  \subfigure[]{\includegraphics[height=0.093\textwidth,width=0.128\linewidth]{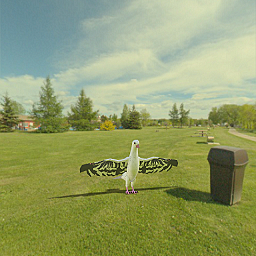}} \\  
\caption{Visual comparison of our method against other start-of-the-art methods on our testing set. (a) 2D foreground object with a illumination condition. (b) Input composite image with a new illumination condition background. (c)-(g) Results produced by ASI3D DoveNet, Intrinsic-Net, ARShadowGAN, and our DIH-GAN respectively. (h) Ground truth.}
\label{fig:visual_comparison}
\end{figure*}
\subsection{Implementation details}
Our DIH-GAN model is implemented by Tensorflow and runs with NVIDIA GeForce GTX 1080Ti GPU. 
We split the 89,898 six-tuples into 71,918 six-tuples for training and 17,980 six-tuples for testing. Note that, there is no crossover between the foreground objects in our training dataset and testing dataset. Our network is trained for 80 epochs, and the resolution of all images for training and testing is $256\times256$. 
The initial learning rate is $10^{-4}$. We set $\beta_{1}$ = 25.0, $\beta_{2}$ = 6.0, $\beta_{3}$ = 0.04, $\beta_{4}$ = 0.5 and adopt Adam optimizer to optimize the DIH-GAN and discriminator.

\section{Experiments}

\subsection{Evaluation Metrics and Experimental Settings}


\noindent{\bfseries Evaluation metrics.} To evaluate the performance of our DIH-GAN, we adopt two commonly-used evaluation metrics including RMSE and SSIM. In addition, we also introduce other two evaluation metrics including fMSE and fSSIM to evaluate the performance on foreground regions. These two metrics are to compute MSE and SSIM values between foreground regions of input and corresponding ground truth. Overall, smaller fMSE, RMSE, and larger fSSIM, SSIM indicate better results.

\noindent{\bfseries Compared methods.} We choose one traditional relighting method ASI3D~\cite{Karsch2014Automatic} with the similar task as ours, and other three deep learning-based methods from the related fields: one shadow generation method
ARShadowGAN~\cite{2020ARShadowGAN}, and two image harmonization methods including DoveNet~\cite{2020DoveNet} and Intrinsic-Net~\cite{guo2021intrinsic}. 
For fair comparison, we re-train ARShadowGan, DoveNet and Intrinsic-Net on our training set, and test them on our testing set for our illumination harmonization task.



\subsection {Comparison with Start-of-the-Art Methods}

\noindent{\bfseries{ Quantitative comparison.}}
Table~\ref{table:Quantitative} reports the quantitative comparison results on our testing set. As can be seen, 
our DIH-GAN achieves the best quantitative results on all these four evaluation metrics.
This is mainly because the traditional methods ASI3D rely on the estimation accuracy of 3D information of objects and scenes. Inaccurate estimation of 3D information often leads to poor results. 
As a deep learning based method, our DIH-GAN does not require complicated 3D information estimation and instead it uses the attention mechanism to enhance the beneficial features for a better result. The best performance of DIH-GAN is mainly attributed to the multi-scale attention mechanism, feature exchange mechanism and adversarial learning, which can better guide the illumination editing of inserted object, refine the features and bridge the illumination gap between inserted object and background environment to obtain results closer to the ground truth.

\noindent{\bfseries{ Visual comparison.}}
We provide some visual comparison results in Figure~\ref{fig:visual_comparison}. 
As we can see, our DIH-GAN not only achieves the illumination transformation of different scenes, but also gains the best visual results with plausible object shadows and harmonious illumination. 
Among these competing methods, ASID3D may estimate  inaccurate information of geometry and illumination of the object and scene. Although ARShadowGAN generates reasonable object shadows, it has weak illumination processing and therefore cannot reasonably edit the object illumination. For Intrinsic-Net and DoveNet, they target at the appearance harmonization of image and can not well address the object shadow (see the 2nd and 3rd rows of Figure~\ref{fig:visual_comparison} (b)(c)). 
In contrast, DIH-GAN is able to achieve the better results with plausible object shadow and harmonious illumination, which mainly because our network makes full use of the collaborative R-Network and I-Network parallel with the multi-scale attention mechanism, the illumination feature exchange mechanism, and the adversarial learning strategy to automatically infer the shadow and illumination generation of the object. 
\begin{table}[] 
\centering

\caption{Quantitative comparison results on our testing set. " $ \uparrow $" indicates the higher the better and " $ \downarrow $" indicates the lower the better.  The best results are marked in \textbf{bold}.}
\vspace{0.2cm}
\resizebox{\linewidth}{!}{
\begin{tabular}{c|c|c|c|cc|c}
\hline
Method &  RMSE $\downarrow$ & fMSE $\downarrow$ & SSIM $\uparrow$ & fSSIM $\uparrow$\\
\hline
ASI3D~\cite{Karsch2014Automatic} &  8.116 & 922.17 &  0.827 & 0.764 \\
DoveNet~\cite{2020DoveNet} &  6.825 &  698.71 &  0.934 & 0.876\\
Intrinsic-Net~\cite{guo2021intrinsic} & 7.493  &  842.18 & 0.921 & 0.803
\\
ARShadowGAN~\cite{2020ARShadowGAN} &  7.043 &  744.06 &  0.928 &  0.812 \\
\hline
DIH-GAN & {\bfseries {6.421}} & {\bfseries { 618.44}} & {\bfseries { 0.957}} & {\bfseries { 0.882}} \\
\hline
\end{tabular}
}
\label{table:Quantitative}
\end{table}
\vspace{-0.2cm}
\subsection{Ablation Study}
We conduct ablation study to evaluate the performance of the proposed multi-scale attention mechanism (MSA), illumination exchange mechanism (IEM) and perceptual loss $\mathcal{L}_{per}$, non-illumination feature loss $\mathcal{L}_{nonillu}$ and adversarial loss $\mathcal{L}_{adv}$.

\begin{figure*}[t]
\centering
\subfigure[]{\includegraphics[width=0.11\linewidth, height=0.068\linewidth]{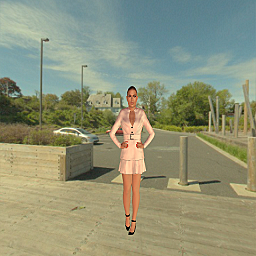}}
\hspace{-4pt}
\subfigure[]{\includegraphics[width=0.11\linewidth, height=0.068\linewidth]{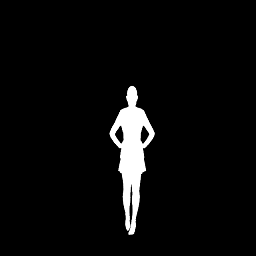}}
\hspace{-4pt}
\subfigure[]{\includegraphics[width=0.11\linewidth, height=0.068\linewidth]{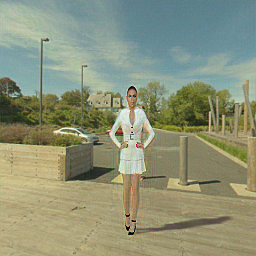}}
\hspace{-4pt}
\subfigure[]{\includegraphics[width=0.11\linewidth, height=0.068\linewidth]{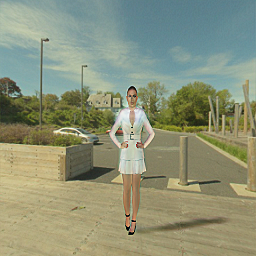}}
\hspace{-4pt}
\subfigure[]{\includegraphics[width=0.11\linewidth, height=0.068\linewidth]{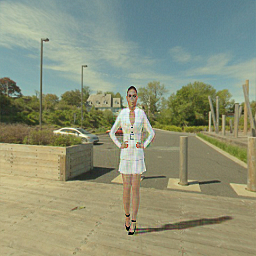}}
\hspace{-4pt}
\subfigure[]{\includegraphics[width=0.11\linewidth, height=0.068\linewidth]{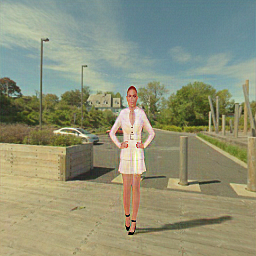}}
\hspace{-4pt}
\subfigure[]{\includegraphics[width=0.11\linewidth, height=0.068\linewidth]{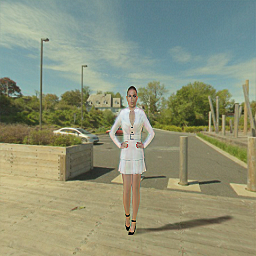}}
\hspace{-4pt}
\subfigure[]{\includegraphics[width=0.11\linewidth, height=0.068\linewidth]{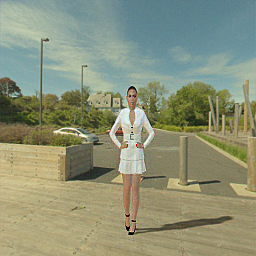}}
\hspace{-4pt}
\subfigure[]{\includegraphics[width=0.11\linewidth, height=0.068\linewidth]{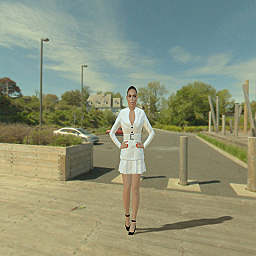}}
\hspace{-4pt}
\caption{Ablation study for our DIH-GAN. (a) Input image without illumination harmonization. (b) mask. (c) Basic + MSA + IEM + $\mathcal{L}_{per}$ + $\mathcal{L}_{nonillu}$. (d) Basic + IEM + $\mathcal{L}_{per}$ + $\mathcal{L}_{nonillu}$ + $\mathcal{L}_{adv}$. (e) Basic + MSA + IEM + $\mathcal{L}_{nonillu}$ + $\mathcal{L}_{adv}$. (f) Basic + MSA + $\mathcal{L}_{per}$ + $\mathcal{L}_{nonillu}$ + $\mathcal{L}_{adv}$. (g) Basic + MSA + IEM + $\mathcal{L}_{adv}$ + $\mathcal{L}_{per}$. (h) DIH-GAN. (i) Ground truth. \textit{Please zoom in to observe the detailed difference.}}
\label{fig:Ablation}
\end{figure*}

The quantitative and visual comparison results are shown in Table~\ref{table:Ablation} and Figure~\ref{fig:Ablation}, respectively. As we can see in Table~\ref{table:Ablation}, our DIH-GAN with all components is able to obtain better results than other methods with one or two components in all three evaluation metrics. By comparing DIH-GAN with “Basic + MSA + $\mathcal{L}_{nonillu}$ + $\mathcal{L}_{adv}$ + IEM”, “Basic + MSA + $\mathcal{L}_{per}$ + $\mathcal{L}_{nonillu}$ + IEM” and “Basic + $\mathcal{L}_{per}$ + $\mathcal{L}_{nonillu}$ + $\mathcal{L}_{adv}$ + IEM” respectively in Table~\ref{table:Ablation}, we find that our proposed multi-scale attention mechanism and the used perceptual loss ($\mathcal{L}_{per}$) and adversial loss ($\mathcal{L}_{adv}$) are all beneficial to our final results.

\begin{table}[] 
\centering
\caption{Ablation study. “Basic” denotes our method without multi-scale attention mechanism (MSA), IEM and the used perceptual loss $\mathcal{L}_{per}$, non-illumination feature loss $\mathcal{L}_{nonillu}$, and adversarial loss $\mathcal{L}_{adv}$. The best results are marked in \textbf{bold}.}
\vspace{0.2cm}
\resizebox{\linewidth}{!}{
\begin{tabular}{c|c|c|cc|c}
\hline
Method &  RMSE $\downarrow$ & SSIM $\uparrow$ &  fSSIM $\uparrow$\\
\hline
Basic & {8.322} & {0.926} & {0.817} \\
Basic + MSA + IEM & {7.074} & {0.948} & {0.831} \\
\hline
Basic + $\mathcal{L}_{adv}$ + IEM & {7.116} & {0.944} & {0.828} \\
Basic + $\mathcal{L}_{per}$ + IEM & {7.148} & {0.935} & {0.822} \\
Basic + $\mathcal{L}_{per}$ + $\mathcal{L}_{nonillu}$ + $\mathcal{L}_{adv}$ + IEM & {7.076} & {0.947} & {0.824} \\
Basic +  MSA + $\mathcal{L}_{adv}$ + $\mathcal{L}_{nonillu}$ + IEM & {6.742} & {0.951} & {0.841} \\
Basic +  MSA + $\mathcal{L}_{per}$ + $\mathcal{L}_{nonillu}$ + IEM & {6.741} & {0.949} & {0.835} \\
Basic + MSA + $\mathcal{L}_{adv}$ + $\mathcal{L}_{per}$ + IEM & {6.522} & {0.953} & {0.858} \\ 
Basic +  MSA + $\mathcal{L}_{per}$ + $\mathcal{L}_{nonillu}$ + $\mathcal{L}_{adv}$ & {7.092} & {0.945} & {0.826} \\
\hline
DIH-GAN & {\bfseries {6.421}} & {\bfseries {0.957}}  & {\bfseries {0.882}}\\
\hline
\end{tabular}
}
\label{table:Ablation}
\vspace{-0.36cm}
\end{table}

From Figure~\ref{fig:Ablation}, we observe that DIH-GAN generates better object shadow and illumination than “Basic + MSA + IEM + $\mathcal{L}_{per}$ + $\mathcal{L}_{nonillu}$” with unnatural result. The “Basic + MSA + IEM + $\mathcal{L}_{per}$ + $\mathcal{L}_{nonillu}$” yields this worse results mainly because the network does not converge, which highlights the advantage of adversarial learning to accelerate network convergence in the task. Another observation is that “Basic + IEM + $\mathcal{L}_{per}$ + $\mathcal{L}_{nonillu}$ + $\mathcal{L}_{adv}$” produces a relatively poor result with coarse object shadow and unnatural illumination compared to DIH-GAN, which demonstrates that our proposed multi-scale attention mechanism can make full use of important features to guide the shadow generation of inserted object and refine the extracted useful features of different scales. In addition, we find that the object with a poor illumination result from (f), mainly because there is no IEM to exchange illumination information. 
Although result (g) is closer to the best one produced by our DIH-GAN, the appearance of the object of DIH-GAN is more refined,
which indicates that the non-illumination feature loss $\mathcal{L}_{nonillu}$ can encourage our network to generate a more accurate illumination harmonization image.

\subsection{Perceptual Study}
\label{subs:PUS}


We conduct a perceptual study as done in \cite{2018Physically} to further evaluate the performance of our method. We choose 100 testing images with various illumination conditions. Among them, 50 images are from the real-world, and the others are generated by our DIH-GAN.


Then we recruited 100 participants from a school campus for subject evaluation. We divide each image into three visual levels: 
(1) {\em Real}: realistic illumination harmonization, (2) {\em Fake}: unrealistic illumination harmonization with artifacts, (3) {\em Uncertain}: uncertain result which they can not make a decision.
52.4\% of real images are judged to be real and the other real images are judged to be fake or uncertain. At the same time, 44.2\% of the images produced by our DIH-GAN are judged to be real, which has almost the same evaluation result as the real images. This illustrates that our network 
can produce high-quality photorealistic results without artifacts.



\subsection{ Discussions}
\begin{figure}[] 
\centering
\includegraphics[width=0.48\textwidth]{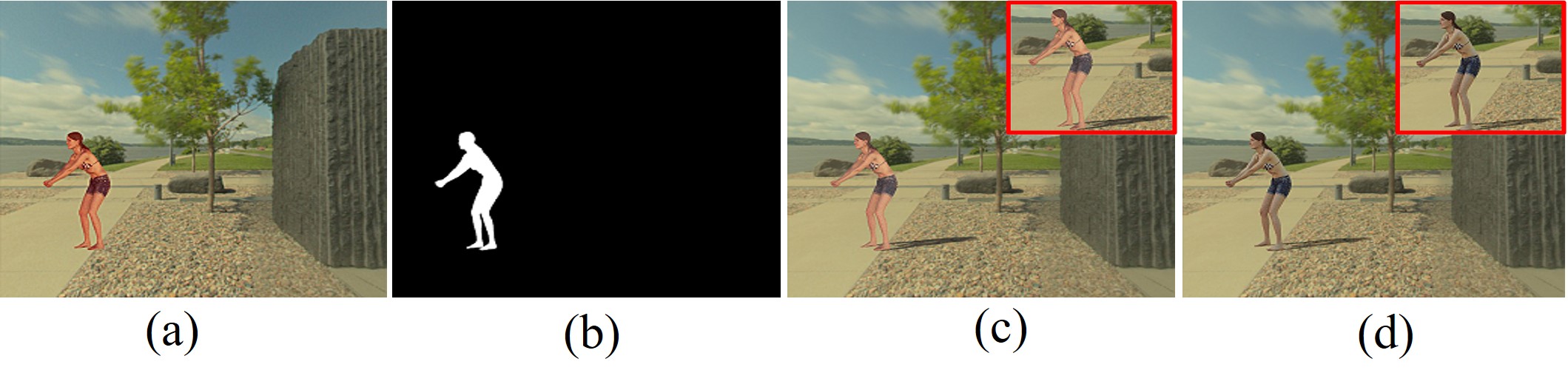}
\caption{The robustness test. (a) Input image. (b) The corresponding input mask. (c)-(d) The illumination harmonization results from ARShadowGAN and the proposed DIH-GAN respectively.}
\label{fig:Robustness}
\end{figure}

\noindent{\bfseries Robustness.}
To verify the robustness of our method, we test our DIH-GAN with new cases outside IH dataset. Specifically, our test cases contain four real-world scenes with different objects, two of which are virtual objects and the other two are real objects from the Internet. 
Note that the ground truths of all testing images are not available. The one visual result is shown in Figure~\ref{fig:Robustness}. From Figure~\ref{fig:Robustness} (c) and (d), we can see that our DIH-GAN not only generates more plausible object shadows than ARShadowGAN, but also successfully achieves the illumination transformation thus to produce the scene illumination harmonization. 

\begin{figure}[] 
\centering
\includegraphics[width=0.48\textwidth]{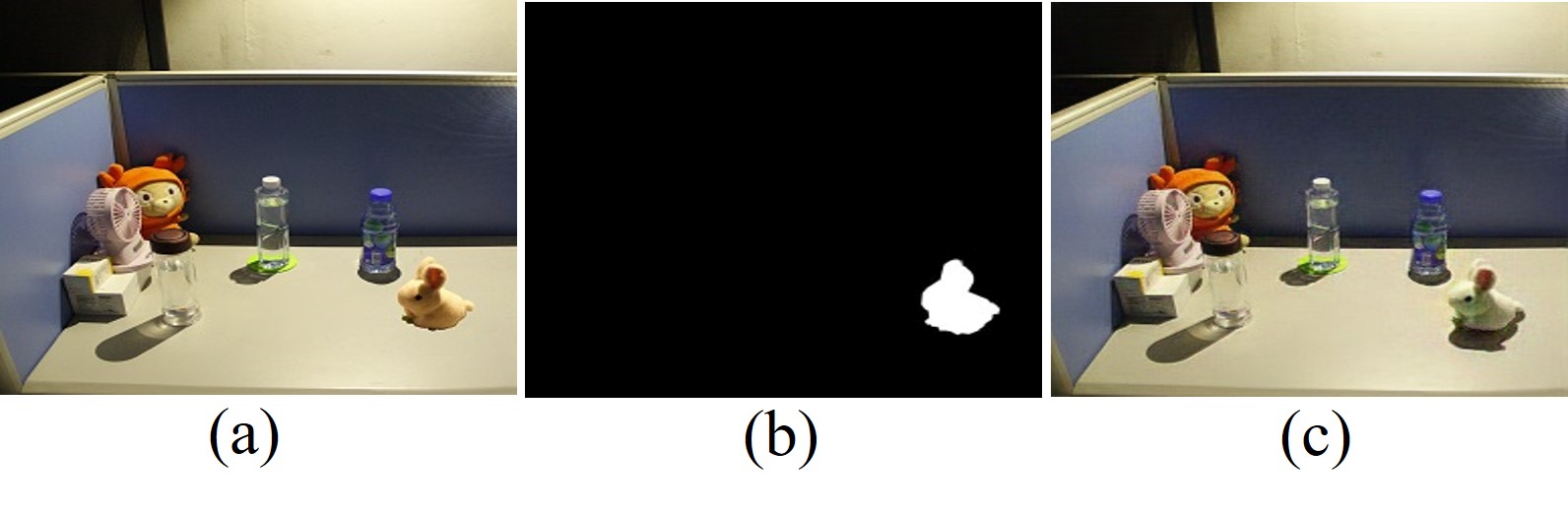}
\caption{Illumination harmonization editing on a real image. From left to right: input image (a), the corresponding input mask (b) and the result (c).}
\label{fig:real_illum_exp}
\end{figure}

\noindent{\textbf{Generalization.}} To verify the generalization ability, we test our DIH-GAN on 200 real-world images. Figure~\ref{fig:real_illum_exp} presents one visual example.
We also adopt the similar perceptual study as done in Section~\ref{subs:PUS} to evaluate the performance of our method.
The evaluation result of subjects is that 62.7\%, 21.3\% and 16.0\% of all results produced by our method are real, fake and uncertain, respectively. It shows that DIH-GAN has a good generalization ability and is able to usually produce high-quality results without visually noticeable artifacts for unseen real-world images.

\noindent{\bfseries Limitations.} Our work has two limitations. First, DIH-GAN mainly focus on the outdoor scenes, and is less able to produce satisfactory results for indoor scenes especially with dark lighting and multiple light sources. Second, our dataset only contains the planar shadows. 
\section{Conclusion and Future Work}
\label{sec:conclusion}
In this work, we have presented a large-scale and high-quality illumination harmonization dataset IH and proposed a novel deep learning-based method DIH-GAN to edit the illumination of the inserted object and generate visually plausible illumination harmonized result without any intermediate inverse rendering process. In the future, we will extend our DIH-GAN and the dataset to address video illumination harmonization.

\noindent{\section*{Acknowledgments}}
This work is partially supported by Prof. Chunxia Xiao's Key Technological Innovation Projects of Hubei Province (2018AAA062) and NSFC (No. 61972298).

\bibliographystyle{ieee_fullname}
\bibliography{egbib}

\end{document}